# MI-VisionShot: Few-shot adaptation of vision-language models for slide-level classification of histopathological images

Pablo Meseguer[1,2,*], Rocío del Amor[1], and Valery Naranjo[1,2]

[1] Instituto Universitario de Investigación e Innovación en Tecnología Centrada en el Ser Humano, HUMAN-tech, Universitat Politècnica de València, València, Spain
[2] valgrAI - Valencian Graduate School and Research Network of Artificial Intelligence (valgrAI), València, Spain
{`pabmees`, `madeam2`, `vnaranjo`}`@upv.es`

**Abstract.** Vision-language supervision has made remarkable strides in learning visual representations from textual guidance. In digital pathology, vision-language models (VLM), pre-trained on curated datasets of histological image-captions, have been adapted to downstream tasks, such as region of interest classification. Zero-shot transfer for slide-level prediction has been formulated by MI-Zero [1], but it exhibits high variability depending on the textual prompts. Inspired by prototypical learning, we propose MI-VisionShot, a training-free adaptation method on top of VLMs to predict slide-level labels in few-shot learning scenarios. Our framework takes advantage of the excellent representation learning of VLM to create prototype-based classifiers under a multiple-instance setting by retrieving the most discriminative patches within each slide. Experimentation through different settings shows the ability of MI-VisionShot to surpass zero-shot transfer with lower variability, even in low-shot scenarios. Code at https://github.com/cvblab/MIVisionShot.

## 1 Introduction

The strategy of pretraining and fine-tuning has become widely explored in computer vision applications, especially with the rise of convolutional neural networks (CNN) and large-scale datasets like ImageNet [2]. This approach involves pretraining models using vast datasets and then fine-tuning them for specific downstream tasks, thus requiring notable computational resources and substantial annotation efforts on the target dataset. This approach changed significantly with the introduction of Constrative Language-Image Pretraining (CLIP) [3]. CLIP suggests learning visual representations by aligning in a contrastive manner image-text pairs gathered from a large-scale web corpus. Unlike the traditional fully-supervised approach for task adaptation, CLIP enables zero-shot image classification by embedding visual categories within text prompts. This innovation holds particular promise for domains with limited data availability, such as the medical field. Vision-language supervision not only facilitates the learning of closely aligned representation spaces between images and language, but also allows transferring vision features to carry out downstream tasks like image classification.

Vision-language models trained on datasets assembled mostly with natural images face challenges in the medical domain due to their inability to encode specific domain knowledge and learn representations in images with a significant domain shift. To overcome this limitation, domain-specific CLIP models were developed by fine-tuning the CLIP weights with multimodal biomedical data covering various organs and diseases [4,5]. Recently, efforts to specialize vision-language models for computational pathology have advanced further. Various data sources, including PubMed articles [6], Twitter [1], and YouTube [7], have been explored to compile extensive datasets of histopathological image-caption pairs used for contrastive pre-training.

The field of computational pathology has experienced a significant transformation in recent years with the increased availability and adoption of digital slide scanning and the rapid advancements in artificial intelligence (AI) research [8]. Among others, task-specific AI tools have been proposed to predict skin cancer [9], perform survival outcome prediction [10] and detect inflammation in ulcerative colitis[11] from whole-slide images (WSI). These methods still initialize the feature extractor with ImageNet pre-trained CNN, which hampers the adaptation performance. The emergence of large models trained in a task-agnostic manner for the histopathological field opens the door to their adaptation to solve different challenges.

The importance of predicting cancer subtypes at the WSI level remains significant in cancer diagnosis. However, the unique nature of gigapixel WSI poses challenges for implementing zero-shot transfer. Recently, MI-Zero [6] has been introduced as a straightforward and intuitive method for transferring the zero-shot capabilities of CLIP-like methods. It consists in combining the similarity of the patch-level image features to the text embeddings. While MI-Zero exhibits impressive performance for slide-level classification tasks in cancer subtyping, it still exhibits a notable reliance on textual prompts.

To overcome this problem, we propose Multiple Instance Vision Shot (MI-VisionShot), an intuitive approach to solving prompt-dependent variability in zero-shot transfer using prototype learning under few-shot scenarios without re-training. This framework works under the multiple instance learning paradigm to handle the gigapixel size of WSI by retrieving the most informative patches according to the textual information. Thus, we obtain a refined global-level embedding that will construct the class prototypes.

## 2    Related work

*A. Adaptation methods for vision-language models*

With the advent of multimodal contrastive pretraining, which involves training models using images and text pairs, zero-shot image classification has become feasible by embedding text prompts containing class names or descriptions of the target classes of the dataset to construct linear classifiers. CLIP [3] has also demonstrated impressive performance in linear probing under few-shot scenarios, promoting the development of module adapters for dual-encoder architectures. Given the pivotal role of text prompts in creating zero-shot classifiers, prompt engineering has emerged. It aims to enhance the discriminative power of each class prompt by incorporating domain-specific expertise but requires prior knowledge of the target task. Therefore, context optimization [12] employs prompt tuning to encode the prompt context into learnable vectors, which are optimized during training.

While prompt tuning focuses solely on optimizing textual inputs, CLIP-Adapter [13] takes a different approach by training lightweight bottleneck layers on top of CLIP-like models. CLIP-Adapter integrates residual connections to preserve pre-trained feature information and mitigate overfitting. Inspired by prototypical networks, TIP-Adapter [14], offers a training-free method for few-shot learning in vision-language models. It utilizes a cache model containing support samples from the training set to encode few-shot knowledge and incorporates zero-shot knowledge from text prompts through weighted integration.

*B. Task-agnostic pretraining and adaptation in digital pathology*

Self-supervised learning has emerged as a powerful approach for model pretraining in computer vision. By leveraging unlabeled data, self-supervised learning enables models to learn meaningful representations from the input data without manual annotations. This is achieved through the design of pretext tasks that encourage the model to extract rich and diverse features that will promote its adaptation to downstream tasks. In digital pathology, self-supervised approaches have been implemented with clustering-guided contrastive learning [15], using hybrid network backbones that combine CNN and transformers [16] or Vision Transformer (ViT) in a teacher-student framework [17].

As mentioned earlier, image-language pre-training aims to learn vision features from natural language supervision. It has been investigated in histopathology image analysis by contrastively aligning paired image and text descriptions. Although all the works follow the same CLIP-inspired approach, models differ in the source of the retrieved dataset. While QUILT [7] and PLIP [1] propose to curate noisy sources like YouTube videos and Twitter posts, CONCH [6] gathers educational sources, and the PubMed Open Access Dataset to retrieve captions form histological images. These works investigate different downstream tasks, such as patch-level classification and cross-modal retrieval. Slide-level image classification has only been explored in [6] by training attention-based mechanisms for patch-level feature aggregation.

## 3    Methodology

An overview of the proposed Multiple Instance-VisionShot (MI-VisionShot) is presented in Figure 1. In the following, the problem formulation and each of the components of the method are detailed.

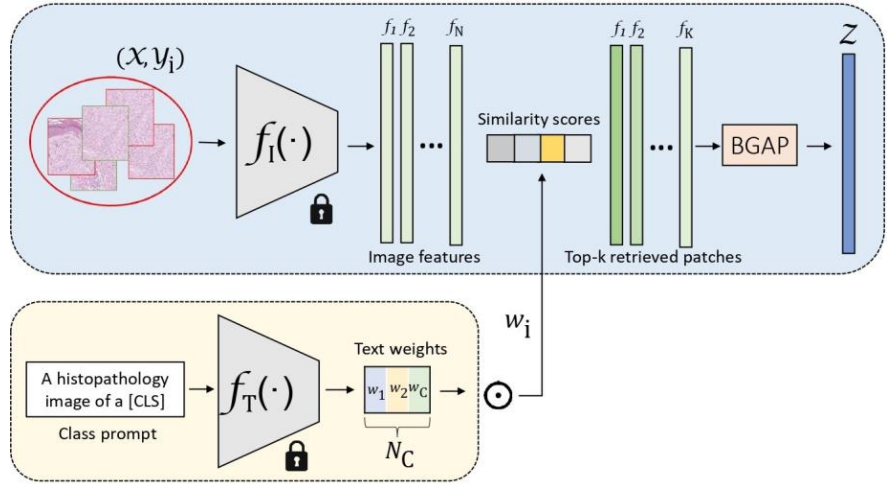

Fig. 1: **Method overview**: We propose MI-VisionShot, a simple but intuitive approach to address prompt dependence in zero-shot transfer for slide-level prediction without retraining. Inspired by prototype learning, MI-VisionShot takes advantage of the aligned latent space between the language and vision representations to obtain discriminative global embedding using the most similar patches to the text embedding of the class.

**Problem Formulation:** Under the multiple instance learning paradigm, instances are grouped into bags $X = \{x_n\}_{n=1}^{N}$, being $N$ the arbitrary number of instances for each bag. In the multi-class scenario, each bag is a member of one of $C$ mutually exclusive classes, such that $Y_c \in \{0, 1\}$. We aim to construct class prototypes under few-shot learning scenarios using a combination of the instance-level projections of the slide that are combined into a global embedding $Z$.

**Vision-language encoding**. Fine-tuning CLIP models using image-caption pairs specific to a certain domain outperforms models trained on more diverse datasets, particularly in clinical settings where there is a significant domain shift and a reliance on expert knowledge. To this end, we select PLIP [1], a vision-language model pre-trained on histopathology image-captions as a baseline for evaluating the proposed adaptation method. Note that our goal is not to compare the VLM itself, but rather to assess the effectiveness of adaptation methods within MIL frameworks without retraining.

Given a target dataset of the downstream task (i.e. cancer subtyping) containing $C$ classes in the form of natural language, the CLIP text encoder embeds the language prompts to create a linear classifier weight matrix $w_T$. The language classifier for zero-shot image classification is generated as follows:

$$w_T = \text{Text-Encoder}(\text{Tokenizer}([H; C_i])) \tag{1}$$

being $H$ a hard template (i.e. *a histopathology image of a [CLS]*) to encode $C_i$ language class names. Given a WSI ($X$) under the multiple instance learning paradigm, the CLIP image-encoder embeds each patch into a lower dimensional manifold and computes $L2$-normalization to obtain patch-level features vectors $f_n$.

$$f_n = \text{Image-Encoder}(\{x_n\}_{n=1}^N) \tag{2}$$

**Prototype learning**. In a few-shot learning paradigm, we aim to construct class prototypes that take advantage of the text information to obtain more discriminative global embeddings, thus contributing to enhancing the separation within classes in the latent space. For a training WSI with a global label $C_i$, we compute the similarity scores ($s_n$) between the text embedding of class $w_i$ and all the feature vectors of the patches.

$$s_n = w_i(\{f_n\}_{n=1}^N) \tag{3}$$

MI-VisionShot retrieves the top-K patches with a greater representation of the class according to the text information. In the following, we perform batch global average pooling (BGAP) of the most informative features ($f_k$) to obtain the refined global embedding of the slide ($Z$).

$$Z = \textbf{BGAP}(f\{argmax(s_k)\}) \tag{4}$$

In the few-shot setting, the prototype of the i-th class $(W_i)$ is finally constructed by averaging the global embeddings of the training samples belonging to the class. Given a bag of the test set, its prediction is assessed by computing the similarities between the BGAP embedding of the instance-level features $(Z')$ and the constructed class prototype ($W$). The prediction is finally assessed by leveraging the largest similarity to the class prototypes (see Equation 5). Note that in the inference step, we cannot obtain the weighed embedding with the textual information as we do not know to which class the test sample belongs.

$$\hat{Y} = \text{argmax}(W^T Z') \tag{5}$$

## 4  Experiments and results

*A. Experimental setting*

To carry out this study, we resort to a public dataset from The Cancer Genome Atlas for subtyping of renal cell carcinoma (RCC). It contains slides of chromophobe RCC (PRCC), clear-cell RCC (CCRCC) and papillary RCC (PRCC). Following [18] for a fair comparison, we used CLAM [19] to segment the tissue of the slide and extract 256x256 patches at equivalent 20x magnification. The overall count of WSI for each of the classes is presented in Table 1.

Table 1: Summary of the TCGA dataset for renal cell carcinoma (RCC) subtyping. The recount of WSI available and the number of patches per WSI is provided for each subtype.

|                  | #WSI | patches/WSI      |
|------------------|------|------------------|
| **Clear cell RCC**   | 513  | 12665 ± 6241     |
| **Chromophobe RCC**  | 119  | 14124 ± 6390     |
| **Papillary RCC**    | 291  | 12429 ± 7305     |
| **All classes**      | 923  | 12779 ± 6634     |

We follow a 5-fold stratified cross-validation to ensure all dataset samples are evaluated. Our framework is evaluated in a few-shot setting with a varying number of $k = \{2, 4, 8, 16\}$ training samples, referring to the number of training samples selected in each setting for constructing the class prototype. For each few-shot scenario, we repeated the experiments with five different seeds to account for the variability inherent in few-shot random sample selection. All models are evaluated in terms of balanced accuracy to equally consider unbalanced classes.

*B. Ablation experiments*

In this section, we present a series of ablation experiments designed to validate the different components of our framework and demonstrate the experimental setup.

**1) Ablation of the number of most informative patches.** WSI can span up to more than 100.000 squared pixels, thus translating a very large number of instances after patching them to follow the MIL paradigm. As noted in Table 1, the average number of patches goes up to more than 12.000 patches per WSI despite not considering background patches with less than 10% of tissue. The main contribution of MI-VisionShot is to select a certain number of patches according to their similarity to the text embedding of the class. To determine the optimal number of patches to select, we evaluate our framework under four different sets of $top-K = \{2, 20, 200, 2000\}$ instances to obtain the aggregated embedding. Table 2 shows the average balanced accuracy and the standard deviation through five seeds for each top-K number of instances. Results show higher performance for $K = 200$ which denotes that using fewer patches is inefficient to obtain enough informative instances, while considering more patches selects tissue sections with less relevant information to determine the diagnosis.

Table 2: Ablation experiments on different numbers of selected instances in MI-VisionShot for slide-level prediction. The best for each number of training samples is highlighted in bold.

|  | k=2 | k=4 | k=8 | k=16 |
|---|---|---|---|---|
| $K = 2$ | 0.705±0.038 | 0.752±0.026 | 0.766±0.013 | 0.780±0.016 |
| $K = 20$ | 0.716±0.019 | 0.744±0.030 | 0.780±0.012 | 0.794±0.007 |
| $K = 200$ | **0.718±0.028** | **0.762±0.014** | **0.783±0.010** | **0.800±0.013** |
| $K = 2000$ | 0.677±0.041 | 0.734±0.027 | 0.776±0.009 | 0.793±0.010 |

**2) Comparative visualization of feature representations.** In the following, we aim to demonstrate how the selection of the most informative patches yields the construction of discriminative prototypes. Inspired in region of interest selection [20], that aim to determine the most relevant instances or features under a MIL paradigm, MI-VisionShot follows a similar approach by retrieving the top-K with a higher similarity score to the text embedding of the class. In Figure 2, we show a 2D T-SNE representation of the BGAP and MI-VisionShot global-level embeddings for all the slides in the dataset. Here, we compare how the text-guided selection of relevant instances leads to a greater separation between classes in the latent space, thus obtaining a silhouette score 3x times larger than with the BGAP embeddings.

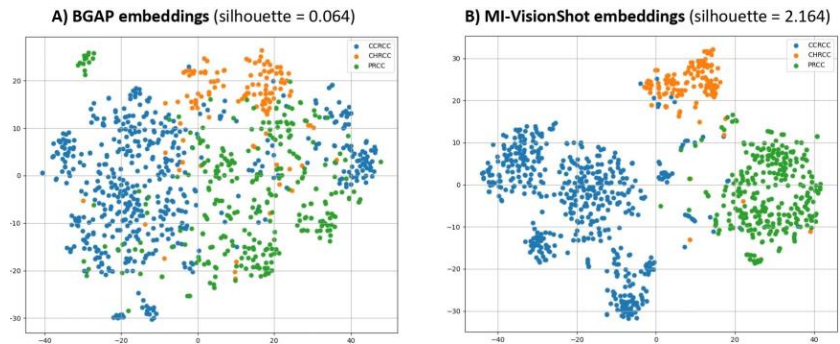

Fig. 2: **T-SNE visualization**: We compare the BGAP embeddings of the vision-language model against the weighted global embeddings obtained with MI-VisionShot. The evaluation utilizes the silhouette score to measure the distinctiveness among clusters representing various classes. A higher silhouette score indicates greater separation between classes.

### C. Comparison with the literature

We initially assess the multi-prompt zero-shot transfer ability of the vision-language model. For that purpose, we follow the experimentation proposed in [18] to create 50 different text classifiers with the textual prompts from the templates and class names. We evaluated MI-Zero with average pooling and without spatial smoothing. We also compare (see Fig. 3) our approach with other training-free adaptation methods based on prototypical learning such as TIP-Adapter [14] and MI-SimpleShot [6] which constructs the class-prototypes as the average of all the available patches.

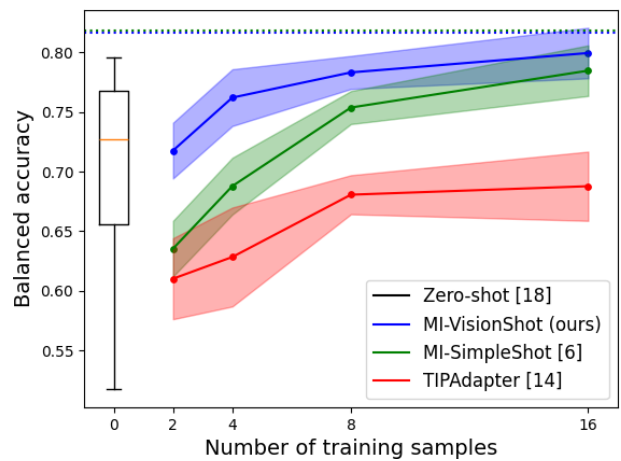

Fig. 3: **Comparison with the literature**: Comparison of MI-VisionShot (ours) with multi-prompted MI-Zero [1], MI-SimpleShot [6] and TIP-Adapter [14]. Solid lines show average accuracy, lighter filled zones standard indicate deviation across seeds and dotted lines represent the upper bound.

First, we evaluated the zero-shot transfer in a multi-prompt configuration which reaches an average of 70.4% balanced accuracy but has an 8.2% standard deviation across different prompts. Results show how the incorporation of prototypical learning in the few-shot paradigm surpasses zero-shot transfer without re-training. MI-VisionShot surpasses MI-SimpleShot by 8.2% and 7.4% in the low-shot learning scenarios ($k = 2$, $k = 4$), respectively. These metrics show that selecting the most representative instances within a bag is even more important in scenarios with fewer data where the prototypes calculated as the average of all samples are not as well defined. TIP-Adapter [14] does not perform in this setting as the construction of the cache knowledge is slightly different than in prototypical learning, where all training samples contribute to the class prototype. It underperforms for slide-level prediction reaching 68.8% balanced accuracy in the higher-shot setting ($k = 16$) in comparison of 79.9% by our proposed approach.

## 5  Conclusions

Vision-language models like CLIP have proven effective in improving AI systems for digital pathology. Task-agnostic learning with in-domain datasets has enabled the adaptation to downstream tasks that were typically handled by task-specific models. Although progress has been made in zero-shot transfer, its performance is still limited for slide-level prediction. Our proposed method, called MI-VisionShot, outperforms existing approaches by constructing prototypes that leverage the most informative patches of each slide. It has been shown to attach prompt dependence in few-shot learning scenarios without retraining and promises the adaptation of VLM to other challenges in cancer subtyping. Additionally, further research should focus on exploiting the alignment between images and natural language in other crucial tasks in digital pathology, such as image captioning.

## 6  Fundings

This work has received funding from the Spanish Ministry of Economy and Competitiveness through projects PID2019-105142RB-C21 (AI4SKIN) and PID2022-140189OB-C21 (ASSIST). The work of Rocío del Amor and Pablo Meseguer has been supported by the Spanish Ministry of Universities under an FPU Grant (FPU20/05263) and valgrAI - Valencian Graduate School and Research Network of Artificial Intelligence, respectively.